\documentclass[11pt]{article}
\usepackage[margin=1in]{geometry}
\usepackage{amsmath,amssymb,amsfonts}
\usepackage[T1]{fontenc}
\usepackage[utf8]{inputenc}
\usepackage{textcomp}
\DeclareUnicodeCharacter{2212}{-}
\usepackage{newtxtext,newtxmath}
\usepackage{graphicx}
\usepackage{multicol}
\usepackage[normalem]{ulem}
\usepackage{booktabs}
\usepackage{multirow}
\usepackage{hyperref}
\usepackage{enumitem}
\usepackage{xcolor}
\usepackage{algorithm}
\usepackage{algorithmic}
\usepackage{natbib}
\usepackage{tikz}
\usetikzlibrary{arrows.meta,positioning,calc}
\usepackage{pifont}
\newcommand{\cmark}{\ding{51}}
\newcommand{\xmark}{\ding{55}}
\usepackage{titling}

\hypersetup{
    colorlinks=true,
    linkcolor=blue!60!black,
    citecolor=blue!60!black,
    urlcolor=blue!60!black
}

\pretitle{%
  \noindent
  {\normalsize\textit{Preprint}}%
  \par\vspace{6pt}%
  \noindent\rule{\linewidth}{1.5pt}%
  \par\vspace{20pt}%
  \begin{center}
  \LARGE
}
\posttitle{%
  \end{center}%
}
\preauthor{\begin{center}\vspace{5pt}\normalsize}
\postauthor{\end{center}}
\predate{\begin{center}\small}
\postdate{\end{center}}

\begin{document}

\title{Preventing Curriculum Collapse in Self-Evolving Reasoning Systems}

\author{\textbf{Vaibhav Mishra}\\[2pt]{\footnotesize\textrm{vaibhavm209625@gmail.com}}}
\date{}
{\setlength{\parskip}{0pt}\maketitle}


\vspace{-15pt}
\begin{abstract}
\noindent
Self-evolving reasoning frameworks let LLMs improve their reasoning capabilities by iteratively generating and solving problems without external supervision, using verifiable rewards. Ideally, such systems are expected to explore a diverse problem space and propose new challenges of high learning value. While prior work has largely focused on solver-side optimisation and verification, recent evidence suggests that self-evolving systems can exhibit \emph{diversity collapse} in posing new problems after just a few iterations, even when surface-level variation is preserved~\cite{li2025rdiverse,zhou2025evolrl}.

We introduce \textbf{Prism}, a question-centric self-evolution method that directly tackles this collapse. Prism defines a persistent diversity signal over an embedding-induced semantic partition of mathematical problems and uses it to encourage balanced exploration of underrepresented regions across iterations. This coverage signal is combined with a Zone-of-Proximal-Development (ZPD) gate to preserve edge-of-solvability difficulty.

Evaluated on seven widely used mathematical reasoning benchmarks against five self-evolving baselines, Prism achieves the highest accuracy on six out of seven tasks, achieving gains of $+3.98$ absolute points over R-Zero on AMC and $+3.68$ on Minerva Math. Prism also generates semantically diverse and challenging questions across iterations, resulting in the construction of the Prism-Math dataset comprising 100k mathematical questions. These results demonstrate that cross-iteration semantic coverage is a high-leverage and under-explored axis for building more capable self-evolving reasoners. We release the code, dataset, and models to facilitate further research.
\end{abstract}
\begin{figure}[h]
\centering
\begin{minipage}{0.49\linewidth}
    \centering
    \includegraphics[width=\linewidth]{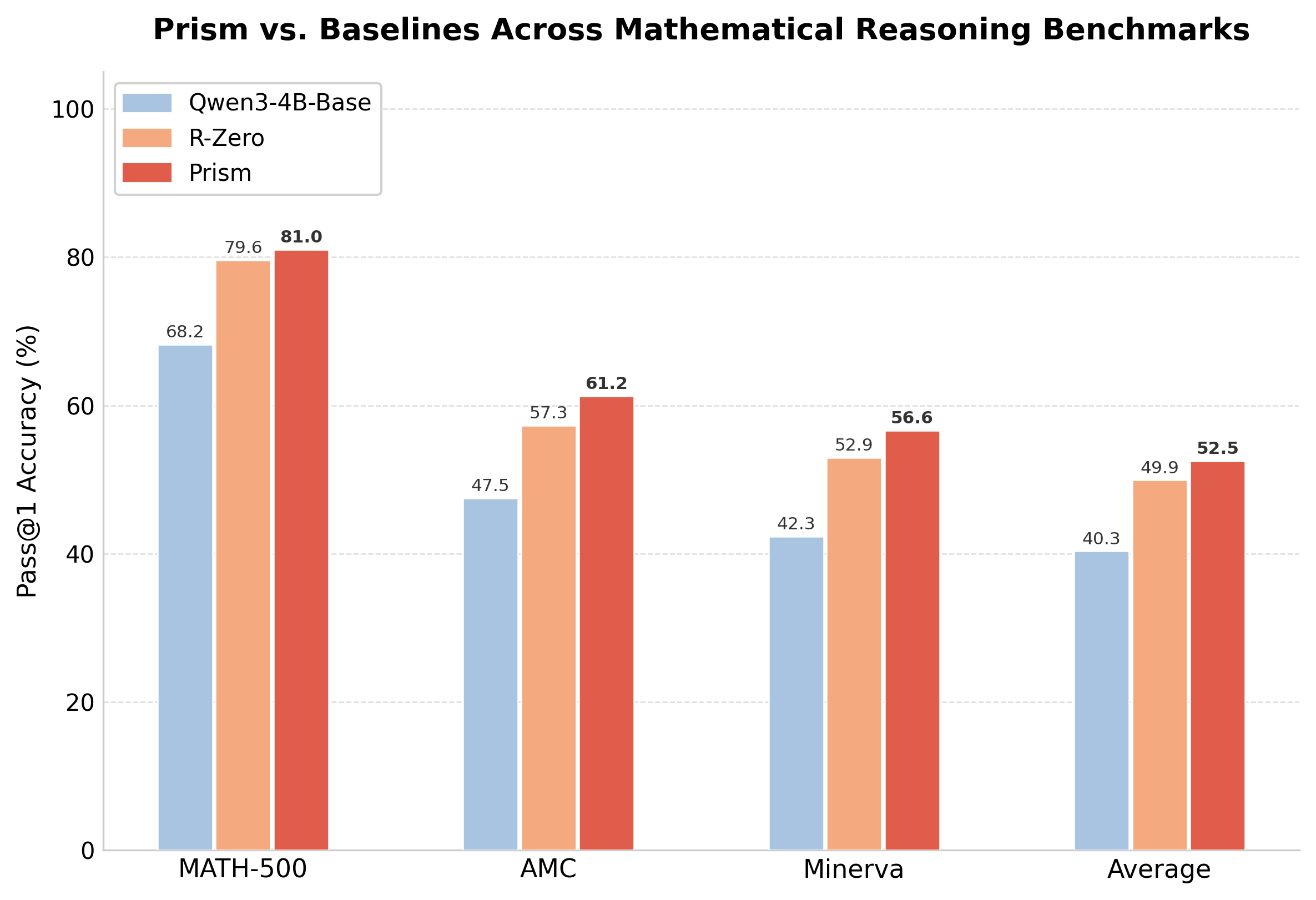}
\end{minipage}\hfill
\begin{minipage}{0.49\linewidth}
    \centering
    \includegraphics[width=\linewidth]{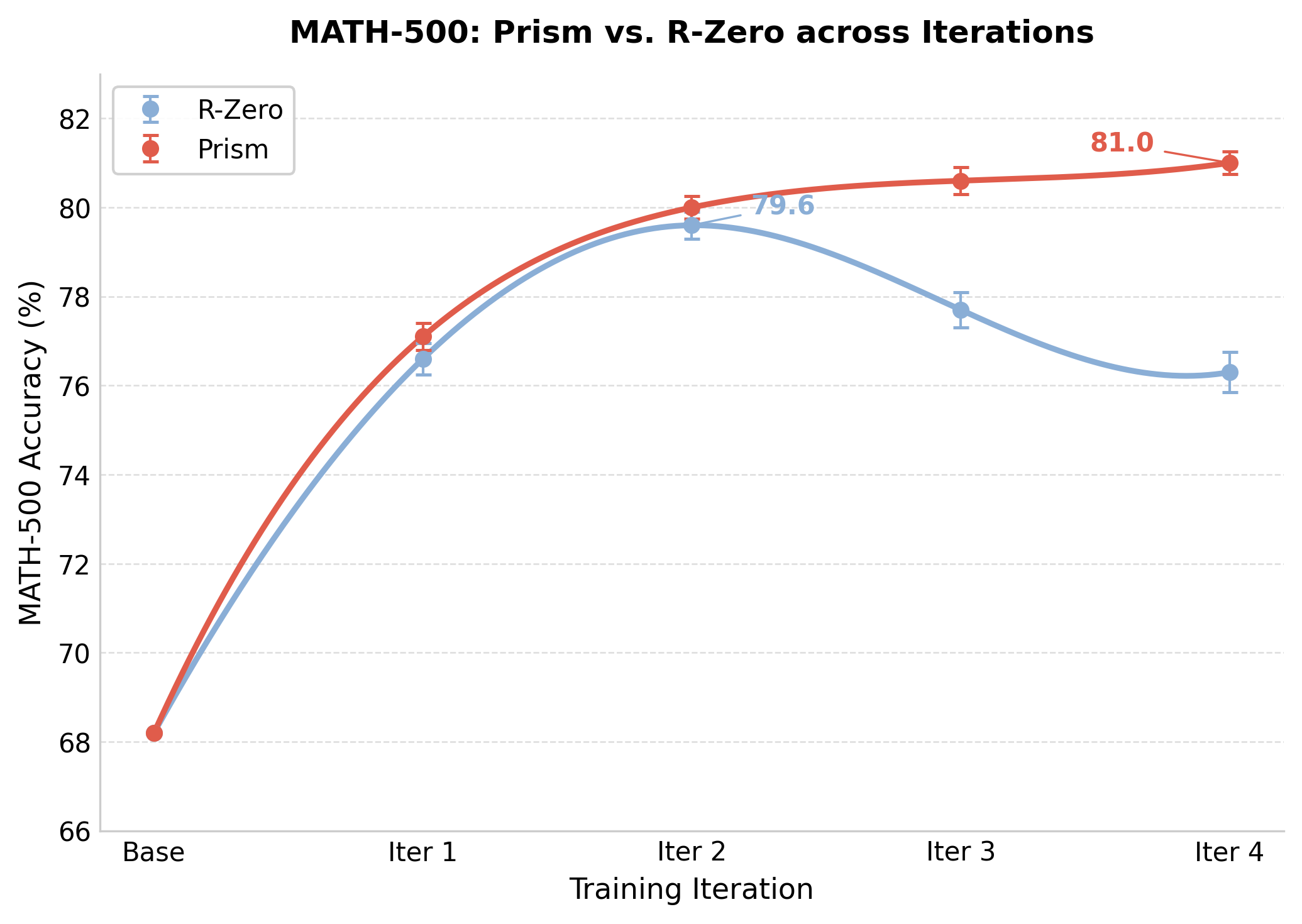}
\end{minipage}
\caption{Prism performance overview. (a) Pass@1 accuracy comparison with baseline across benchmarks. (b) Math-500 test accuracy across iterations.}
\label{fig:main_results_training}
\end{figure}

\vspace{-6pt}
\begin{center}
\small
\textbf{Resources:}\quad
\href{https://github.com/vaibhavmishra1/Prism}{\texttt{Code}}\quad$\vert$\quad
\href{https://huggingface.co/vibhuiitj/Prism-Questioner}{\texttt{Questioner Model}}\quad$\vert$\quad
\href{https://huggingface.co/vibhuiitj/Prism-Solver}{\texttt{Solver Model}}\quad$\vert$\quad
\href{https://huggingface.co/datasets/vibhuiitj/Prism-Math}{\texttt{Prism-Math Dataset}}
\end{center}
\vspace{-6pt}

\section{Introduction}

Reinforcement learning with verifiable rewards (RLVR) has substantially advanced mathematical reasoning in large language models (LLMs)~\cite{guo2025deepseekr1,shao2024deepseekmath}.
In parallel, a growing line of work has explored \emph{self-evolution}, in which training systems improve 
by generating and solving their own tasks, reducing dependence on curated supervision and enabling 
scalable self-loop learning~\cite{silver2017alphazero,chen2024spin,huang2025rzero,yue2026drzero,
chen2025selfevolvingcurriculum,kuba2025languageselfplay}.
In this paradigm, a \textbf{Questioner} proposes problems and a \textbf{Solver} learns from them under verifiable reward signals such as majority voting or self-verification~\cite{huang2025rzero,zhou2025evolrl,shao2025deepseekmathv2}.

However, training on self-generated data can cause concept drift or entropy collapse, degrading curriculum diversity and inducing repetition or forgetting over time~\cite{shumailov2023curse,dohmatob2024modelcollapse}.
Difficulty calibration (ZPD-style gating) provides a \emph{local} learnability signal but does not create pressure to visit underrepresented regions of the problem space.
R-Zero employs an online BLEU-based repetition penalty to discourage lexical similarity 
within a batch.
However, lexical signals fail to detect semantic equivalence, operate only locally within a batch, and 
lack cross-iteration memory.
R-Few~\cite{yu2025rfew} stabilises co-evolution by injecting a small pool of human-labelled anchor examples as in-context guidance, but this diversity is externally induced and requires ongoing access to labelled data.

Recent work has begun diagnosing related issues.
R-Diverse~\cite{li2025rdiverse} identified the diversity illusion in self-play training, while 
Evol-RL~\cite{zhou2025evolrl} studied solution-space entropy collapse under majority-driven selection 
and proposed novelty-aware rewards to preserve variation.
Absolute Zero~\cite{zhao2025absolutezero} eliminates external data entirely by having a single model propose and solve code-reasoning tasks validated by a code executor, achieving strong cross-domain transfer to mathematics.
SPICE~\cite{chen2025spice} leverages corpus-grounded self-play in which the model trains on self-generated questions conditioned on retrieved documents, improving reasoning through environment interaction.
Other directions improve solver reasoning via data scaling and self-improvement pipelines~\cite
{zeng2024skyworkmath,yang2024qwen25mathtech,pei2025scalediff,wei2025learningtoposeproblems,
lu2024mathgenie}.
However, much of this progress remains \emph{solver-centric}.
In contrast, we argue that the bottleneck lies upstream, in the \emph{semantic coverage} and \emph{difficulty calibration} of the generated curriculum.
In a co-evolutionary loop, the questioner’s output defines the solver’s training distribution; a solver 
cannot learn strategies it is never prompted to exercise.
Therefore, sustained capability growth requires explicit control over cross-iteration coverage in the 
question generation process.

We introduce \textbf{Prism}, a question-centric, self-evolution framework that explicitly regularises \emph{cross-iteration semantic coverage} during problem generation.
Prism defines a persistent coverage signal over an embedding-partitioned mathematical space and rewards questions from underexplored regions, while a ZPD gate preserves edge-of-solvability difficulty.
In addition, Prism initialises each new Questioner from the latest Solver rather than the previous Questioner, reducing capability lag and providing a fresh starting point that can support more unbiased exploration when combined with an explicit question-side diversity objective.
Beyond improving Solver performance, Prism also serves as an effective generator of diverse and 
high-difficulty mathematical problems. 
By jointly enforcing edge-of-solvability constraints and cross-iteration semantic coverage, it produces 
curricula that span broad reasoning domains while reaching competition-level hardness, enabling 
the construction of high-quality mathematical datasets with limited manual curation.

Across seven mathematical reasoning benchmarks, Prism outperforms five self-evolving baselines on six of seven tasks, with especially large gains on harder evaluations (e.g., AIME~2024, Minerva Math, and AMC).
These results identify cross-iteration semantic coverage as a high-leverage axis for advancing self-evolving reasoners.
We summarise our key contributions as follows:
\begin{enumerate}[leftmargin=*,itemsep=2pt]
    \item We \textbf{identify the solver-centric bias} in the self-evolving LLM literature and argue that the overlooked lever for improvement is the \emph{questioner}, whose semantic diversity and quality directly determine the solver's performance ceiling.
    \item We \textbf{diagnose curriculum collapse} as the primary failure mode of self-evolving reasoning LLMs and show that existing lexical diversity penalties are insufficient to prevent it.
    \item We propose \textbf{Prism}, a question-centric self-evolution method that treats question generation as a first-class optimisation target within a self-evolving loop, comprising (a) a \textbf{semantic cluster-diversity reward} that provides persistent coverage pressure over a pre-computed semantic partition, and (b) a \textbf{Solver-initialised Questioner} strategy that reduces capability lag and mitigates progressive distributional narrowing by re-deriving the Questioner from the Solver at each iteration.
    \item We provide \textbf{empirical evidence} across seven benchmarks that improved semantic curriculum coverage is associated with stronger generalisation in solver capabilities.
    \item We release \textbf{Prism-Math}, a dataset of ${\sim}$100K semantically diverse, difficulty-calibrated synthetic mathematical questions generated by the Prism questioner, spanning diverse mathematical topics, providing a resource for training and evaluating mathematical reasoning models.
\end{enumerate}

\section{Related Work}

\paragraph{Reinforcement learning for reasoning.}
Outcome-based RL has been effective for mathematical reasoning.
DeepSeekMath~\cite{shao2024deepseekmath} demonstrated that GRPO enables stable optimisation of verifiable reasoning skills, while DeepSeek-R1~\cite{guo2025deepseekr1} showed that strong reasoning can emerge through pure RL without supervised fine-tuning.
These advances establish RLVR as a practical training paradigm for both solvers and question generators in closed-loop systems.

\paragraph{Self-evolving curricula and diversity collapse.}
Inspired by AlphaZero-style self-play~\cite{silver2017alphazero}, self-play fine-tuning (SPIN)~\cite{chen2024spin} demonstrated that a model can improve by learning from its own generated outputs without external labels, a precursor to the fully closed-loop frameworks that followed.
Recent frameworks extend this idea by training a questioner and solver pair without any external data~\cite{huang2025rzero,yue2026drzero,chen2025selfevolvingcurriculum,kuba2025languageselfplay}.
R-Diverse~\cite{li2025rdiverse} formalised the \emph{diversity illusion}, where surface variation masks semantic repetition, and Evol-RL~\cite{zhou2025evolrl} identified entropy collapse focused on the solution space.
R-Few~\cite{yu2025rfew} takes a different approach by injecting a small pool of human-labelled anchor examples as in-context prompts to guide the Questioner; this delays the performance plateau but introduces a systematic \emph{question-side bias} toward the human-curated topic distribution and requires continuous access to labelled data.
Absolute Zero~\cite{zhao2025absolutezero} removes external data entirely by having a single model propose and solve code-reasoning tasks grounded in a code executor that serves as a unified source of verifiable reward.
By casting task generation as code synthesis validated by execution, Absolute Zero achieves strong cross-domain transfer to mathematical reasoning despite training exclusively on self-proposed code tasks.
SPICE~\cite{chen2025spice} introduces corpus-grounded self-play in which the model generates questions conditioned on retrieved passages and trains on its own outputs, improving reasoning through interaction with a textual environment rather than explicit diversity signals.

\paragraph{Curriculum learning, problem synthesis, and quality-diversity.}
Curriculum learning orders training by difficulty~\cite{bengio2009curriculum}; quality-diversity methods such as MAP-Elites~\cite{pugh2016mapelites} maintain solution archives across behavioural niches to prevent mode collapse.
A complementary line of work generates or scales high-quality math problems~\cite{lu2024mathgenie,pei2025scalediff,wei2025learningtoposeproblems,zeng2024skyworkmath,yang2024qwen25mathtech,shao2025deepseekmathv2,yuan2025naturalreasoning,liu2025saturn}.
Liang et al.~\cite{liang2025beyondpass} show that combining self-play with variational problem synthesis can sustain RLVR beyond the single-iteration pass@1 bottleneck.
\section{Preliminaries}
\label{sec:preliminaries}

We introduce notation and formalise the building blocks of Prism.

\paragraph{Notation.}
Let $\mathcal{Q}$ denote the space of mathematical problems and $\mathcal{A}$ the space of candidate solutions.
A \emph{questioner} $Q_\theta$ parameterised by $\theta$ defines a generative policy $\pi^Q_\theta$ over $\mathcal{Q}$, and a \emph{solver} $S_\phi$ parameterised by $\phi$ defines a conditional policy $\pi^S_\phi(\cdot\mid q)$ over $\mathcal{A}$.
Training alternates between the two across $T$ co-evolution iterations indexed by $t$~\cite{huang2025rzero}.

\paragraph{Solvability scoring, policy optimisation, and ZPD gating.}
Given a question $q$ and the current solver $S_\phi$, we draw $n$ independent roll-outs $\{a_i\}_{i=1}^{n}\!\sim\!\pi^S_\phi(\cdot\mid q)$ and apply a verifier $\mathsf{V}\!:\mathcal{Q}\!\times\!\mathcal{A}\!\to\!\Delta(\{0,1\})$.
The \emph{solvability score}
\begin{equation}
    p(q) \;=\; \frac{1}{n}\sum_{i=1}^{n}\mathsf{V}(q,a_i)\;\in\;[0,1]
    \label{eq:majority_rate}
\end{equation}
provides a label-free, continuous estimate of problem difficulty relative to the current solver~\cite{huang2025rzero,zhou2025evolrl}.

In this work, both agents are optimised with Group Relative Policy Optimisation (GRPO)~\cite{shao2024deepseekmath}, a variance-reduced policy-gradient method suited to sparse, outcome-based rewards.
For a prompt $x$, a group of $G$ completions $\{y_g\}_{g=1}^{G}\!\sim\!\pi_\psi(\cdot\mid x)$ is sampled and scored, yielding rewards $\{r_g\}$.
Advantages are computed relative to the within-group mean:
\begin{equation}
    A_g \;=\; r_g \;-\; \frac{1}{G}\sum_{j=1}^{G}r_j\,,
    \label{eq:grpo_advantage}
\end{equation}
and the policy is updated by maximising a clipped surrogate objective with KL regularisation:
\begin{equation}
    \mathcal{J}(\psi)
    \;=\;
    \mathbb{E}_{x}\!\left[
        \frac{1}{G}\sum_{g=1}^{G}
        \min\!\Big(
            \rho_g\,A_g,\;
            \mathrm{clip}(\rho_g,1\!-\!\epsilon,1\!+\!\epsilon)\,A_g
        \Big)
    \right]
    -\;\beta\,\mathrm{KL}\!\bigl(\pi_\psi\|\pi_{\mathrm{ref}}\bigr),
    \label{eq:grpo_obj}
\end{equation}
where $\rho_g = \pi_\psi(y_g\!\mid\!x)/\pi_{\mathrm{ref}}(y_g\!\mid\!x)$ is the importance ratio, $\epsilon$ the clipping threshold, and $\beta$ the KL penalty coefficient.
The group baseline in Equation~\ref{eq:grpo_advantage} eliminates the need for a learned value function, which is particularly advantageous when reward signals are binary or near-binary~\cite{guo2025deepseekr1}.

Finally, self-evolving questioners should generate problems that are neither trivially solvable nor intractable (i.e., problems at the \emph{edge of solvability}~\cite{huang2025rzero}).
We formalise this via a piecewise-linear ZPD gate $g:[0,1]\to[0,1]$:
\begin{equation}
    g(p)
    \;=\;
    \max\!\Bigl(0,\;1-\tfrac{|p-p^\star|}{\Delta}\Bigr)
    \;\cdot\;
    \mathbf{1}\!\bigl[p\in[p_{\min},\,p_{\max}]\bigr],
    \label{eq:zpd_gate}
\end{equation}
which zeroes out reward for questions that are too easy ($p>p_{\max}$) or too hard ($p<p_{\min}$), concentrating reward near target solvability $p^\star$.
In Prism we set $p_{\min}=0.5$ and $p_{\max}=0.9$ throughout.

\paragraph{Self-evolving questioner--solver loop.}
Most self-evolving reasoning frameworks can be expressed as an alternating optimisation over a questioner $Q_{\theta_t}$ and solver $S_{\phi_t}$ at iteration $t$. The questioner induces a sampling distribution $\pi^Q_{\theta_t}$ from which we draw a batch $\{q_g\}_{g=1}^{G} \sim \pi^Q_{\theta_t}$. For each $q_g$, the solver draws $n$ roll-outs $\{a_{g,i}\}_{i=1}^{n} \sim \pi^S_{\phi_t}(\cdot \mid q_g)$. A verifier $\mathsf{V}$ (e.g., a deterministic rule, a symbolic checker, or a separate LLM judge) assigns a binary score to each roll-out, and the empirical solvability estimate
\begin{equation*}
    \hat{p}(q_g) \;=\; \frac{1}{n}\sum_{i=1}^{n}\mathsf{V}(q_g,a_{g,i})
\end{equation*}
(see Equation~\ref{eq:majority_rate}) reflects difficulty. The questioner is then updated by an RLVR algorithm (e.g., GRPO) using rewards of the form $r(q_g)=\tilde{r}(\hat{p}(q_g),q_g)$, typically incorporating a ZPD gate $g(\hat{p}(q_g))$ (Equation~\ref{eq:zpd_gate}). The solver is trained on questions that pass the difficulty filter, using policy optimisation (e.g., GRPO) with outcome-based rewards or pseudo-labels derived from the same verifier $\mathsf{V}$.

\paragraph{Semantic partition and EMA coverage tracking.}
To move beyond surface-level diversity, Prism operates over a fixed semantic partition of $\mathcal{Q}$.
Let $f:\mathcal{Q}\to\mathbb{S}^{d-1}$ be a pre-trained text embedding model and $\boldsymbol{M}=\{\boldsymbol{\mu}_k\}_{k=1}^{K}$ be centroids obtained by running $K$-Means on the MATH training set (${\approx}$12.5K problems~\cite{hendrycks2021math}).
Each question is assigned to its nearest centroid:
\begin{equation}
    c(q)
    \;=\;
    \arg\max_{k\in[K]}\;\bigl\langle f(q),\,\boldsymbol{\mu}_k\bigr\rangle.
    \label{eq:cluster_assign}
\end{equation}
Both $f(q)$ and $\boldsymbol{\mu}_k$ are $L_2$-normalised, so the inner product corresponds to cosine similarity.
The centroids are computed once offline and remain fixed, defining a static coordinate system over embedding-defined regions of the problem space.
To track how frequently each region has been visited, Prism maintains a count vector $\mathbf{n}\in\mathbb{R}_{\ge 0}^{K}$ updated after each batch via exponential moving average:
\begin{equation}
    n_k
    \;\leftarrow\;
    \gamma\,n_k \;+\; (1-\gamma)\,m_k,
    \qquad k=1,\dots,K,
    \label{eq:ema_update}
\end{equation}
with decay $\gamma\in(0,1)$.
Here $m_k$ denotes the number of generated questions in the current batch assigned to cluster $k$ (so counts are updated per-question, with decay applied once per batch).
Equivalently, one may implement this update by first applying a single decay $n_k \leftarrow \gamma n_k$ for all $k$ and then incrementing $n_{c(q)} \leftarrow n_{c(q)} + (1-\gamma)$ once per generated question $q$ in the batch (as in Algorithm~\ref{alg:prism}); this yields $n_k \leftarrow \gamma n_k + (1-\gamma)m_k$.
This provides \emph{cross-iteration memory}, since over-sampled clusters retain elevated counts across co-evolution rounds, a property absent from batch-local diversity penalties.

\paragraph{Coverage-aware questioner objective.}
Prism composes the ZPD gate with a multiplicative coverage bonus.
For a question $q$ with solvability $p(q)$ and cluster assignment $c(q)$, define a \emph{rarity bonus}
\begin{equation}
    d(q)
    \;=\;
    \exp\!\Bigl(-\frac{n_{c(q)}}{\bar{n}}\Bigr),
    \qquad
    \bar{n}=\frac{1}{K}\sum_{k=1}^{K}n_k\,,
    \label{eq:rarity_bonus}
\end{equation}
which is monotone decreasing in the normalised visit count of $q$'s cluster.
Normalising by $\bar{n}$ makes the bonus depend on \emph{relative} visitation: if all counts scale by a constant (e.g., due to longer training), the ratio $n_{c(q)}/\bar{n}$ is unchanged.
The full questioner reward is
\begin{equation}
    r(q)
    \;=\;
    \underbrace{g\!\bigl(p(q)\bigr)}_{\text{ZPD gate (quality)}}
    \;\cdot\;
    \underbrace{\bigl(1+\lambda\,d(q)\bigr)}_{\text{coverage bonus (diversity)}},
    \label{eq:prism_reward}
\end{equation}
where $\lambda>0$ controls the strength of coverage regularisation.
The multiplicative form ensures that coverage pressure \emph{amplifies} the ZPD signal rather than competing with it. A question in a rare cluster receives a larger reward than an equally solvable question in a common cluster, but an unsolvable or trivial question receives zero reward regardless of rarity.
This structure encourages a quality-diversity trade-off analogous to MAP-Elites~\cite{pugh2016mapelites}, with the ZPD gate as the quality filter and the rarity bonus as the diversity objective.

\section{Method}
\label{sec:method}

Building on the co-evolutionary framework described in Section~\ref{sec:preliminaries}, Prism modifies the Questioner's reward function and initialisation strategy while leaving Solver training unchanged.
R-Zero's difficulty-targeting signals (the uncertainty reward and BLEU-based repetition penalty) are \emph{necessary but not sufficient}. They ensure generated questions sit at an appropriate difficulty level, but exert no pressure to cover the breadth of the mathematical problem space.
Without such pressure, the Questioner converges to a narrow set of familiar problem types that reliably achieve high reward, and the curriculum gradually collapses.
We hypothesise that persistent coverage regularisation improves optimisation by reducing question-side reward overfitting to a small set of high-reward templates and by broadening the solver's training distribution across distinct embedding-defined regions. This yields more diverse gradient signal over the solver's reasoning skills, improving generalisation on out-of-distribution benchmarks and delaying the self-reinforcing feedback loop that drives curriculum collapse.

\subsection{Offline Cluster Space Construction}
\label{sec:cluster_construction}

Before any training begins, we construct a semantic partition of the mathematical problem space.
This partition serves as a \emph{fixed coordinate system} over mathematical problems, against which we can measure and incentivise coverage:

\begin{enumerate}[leftmargin=*,itemsep=2pt]
    \item \textbf{Embedding.} We use the training split of the MATH dataset~\cite{hendrycks2021math} ($\mathcal{D}$, ${\approx}$12.5K problems) as the reference corpus. Every question in $\mathcal{D}$ is embedded using \texttt{Qwen3-Embedding-0.6B}, producing $L_2$-normalised vectors $\mathbf{e}_i \in \mathbb{R}^d$ ($d{=}1024$).
    \item \textbf{Clustering.} $K$-Means is applied to $\{\mathbf{e}_i\}$ with $K{=}128$ clusters. The resulting centroids $\{\boldsymbol{\mu}_1, \ldots, \boldsymbol{\mu}_K\}$ are $L_2$-normalised and saved as a static artifact.
    \item \textbf{Initialisation.} A per-cluster visit-count vector $\mathbf{n} \in \mathbb{R}^K$ is initialised uniformly: $n_k = \alpha$ for all $k$, where $\alpha$ is a smoothing constant.
\end{enumerate}
This is a one-time offline step; the centroids are never updated during training.
Note that this semantic partition is constructed from the MATH training split, so Prism is not a fully data-free self-evolution method in the strict sense.
The MATH training set provides broad coverage of competition-level mathematical topics, making it a natural coordinate system for mathematical question diversity.
The 128 clusters correspond to embedding-defined regions that often (but not always) align with familiar mathematical topics such as number theory, combinatorics, various geometry sub-types, algebraic manipulation, calculus, and probability, providing a coarse partition of the mathematical landscape.
Crucially, this partition is defined over the \emph{semantic embedding space}, not over surface lexical features, enabling it to detect redundancy that BLEU-based methods fundamentally cannot.

\begin{figure}[h]
\centering
\includegraphics[width=0.9\linewidth]{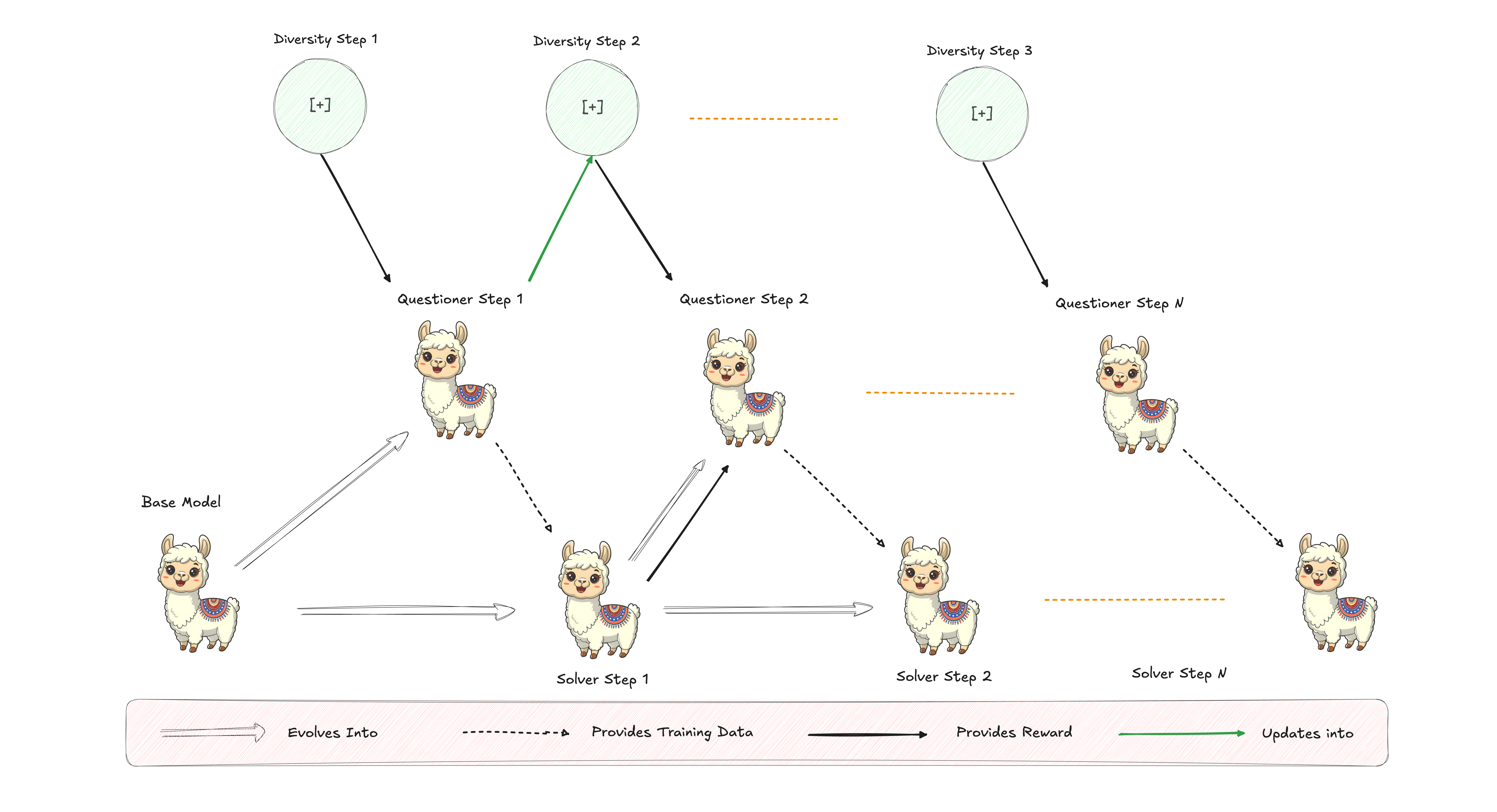}
\caption{Iteration-wise progression of the Prism self-evolution loop.}
\label{fig:iterations}
\end{figure}

\subsection{Cluster-Diversity Reward}

At each Questioner training step, after majority-vote scoring of generated questions, the reward is computed as follows.

\paragraph{Step 1: Cluster assignment.}
Each generated question $q$ that passes the majority-vote threshold ($p(q) \geq 0.5$) is embedded using \texttt{Qwen3-Embedding-0.6B} and assigned to its nearest centroid from the pre-built cluster space (Section~\ref{sec:cluster_construction}):
\begin{equation}
    c(q) = \arg\max_{k \in [K]}\; \mathbf{e}_q^\top \boldsymbol{\mu}_k.
\end{equation}

\paragraph{Step 2: Rarity reward.}
For a question assigned to cluster $c$, the rarity reward is
\begin{equation}
    r_{\text{rarity}}(c) \;=\; \exp\!\left(-\frac{n_c}{\bar{n}}\right), \qquad \bar{n} = \frac{1}{K}\sum_{k=1}^{K} n_k,
    \label{eq:rarity}
\end{equation}
where $n_c$ is the current EMA-smoothed visit count for cluster $c$.
This exponential mapping yields a smooth, bounded bonus in which under-visited clusters ($n_c \ll \bar{n}$) receive a larger reward and frequently visited clusters receive a smaller one.
While other monotone decreasing functions of $n_c$ are possible, we use the exponential form for its simplicity and stable scaling when combined multiplicatively with the ZPD gate (Equation~\ref{eq:final_reward}).

\paragraph{Step 3: ZPD-gated final reward.}
The diversity bonus is modulated by a ZPD gate to ensure that exploration does not compromise solvability:
\begin{equation}
    \text{zpd}(p) \;=\; \max\!\left(0,\; 1 - \frac{|p - 0.75|}{0.4}\right), \quad p \in [0.5, 0.9],
    \label{eq:zpd}
\end{equation}
\begin{equation}
    r_{\text{final}}(q) \;=\; \text{zpd}\bigl(p(q)\bigr) \cdot \Bigl(1 + \lambda \cdot r_{\text{rarity}}\bigl(c(q)\bigr)\Bigr),
    \label{eq:final_reward}
\end{equation}
with $\lambda = 5.0$.
The ZPD peaks at $p = 0.75$, reflecting an asymmetric preference for questions that the Solver can solve 75\% of the time over those at the 50/50 boundary, since the former provide more stable gradient signal while remaining informative.
Questions outside the $[0.5, 0.9]$ range receive zero reward regardless of diversity, ensuring that the rarity bonus cannot incentivise the generation of unsolvable or trivial problems.

The multiplicative structure of Equation~\ref{eq:final_reward} ensures that the rarity reward \emph{amplifies} the ZPD signal rather than adding to it.
A question in a rare cluster receives a higher reward than an equally-solvable question in a common cluster, but an unsolvable question receives zero reward regardless of cluster rarity.
This ensures that diversity and quality are coupled rather than traded off, mirroring the quality-diversity principle advocated by MAP-Elites~\cite{pugh2016mapelites} and related work in evolutionary computation.

\paragraph{Step 4: Count update.}
After each batch, cluster counts are updated with exponential moving average (EMA) decay:
\begin{equation}
    n_c \;\leftarrow\; \gamma \cdot n_c + (1 - \gamma)\,m_c,
    \label{eq:ema}
\end{equation}
with $\gamma = 0.99$.
Here $m_c$ denotes the number of questions in the batch assigned to cluster $c$ (equivalently: decay counts once per batch, then add $(1-\gamma)$ once per question assigned to $c$).
The decay ensures responsiveness to distribution shifts, so clusters that were popular early in training can become ``rare'' again if the Questioner's distribution drifts, preventing the count vector from becoming a stale artifact.

\paragraph{Cross-iteration warm-start.}
At iteration $t > 1$, we warm-start the cluster frequency vector $\mathbf{n}$ from the distribution observed at the end of iteration $t{-}1$.
This provides \emph{cross-iteration memory}, a feature absent from R-Zero's batch-local BLEU penalty. This ensures that diversity pressure compounds rather than resets across iterations, and prevents the Questioner from redundantly re-exploring already-saturated clusters.

\subsection{Questioner Training}

In R-Zero, the Questioner at iteration $t$ is initialised from the previous Questioner $Q_{t-1}$.
In Prism, we instead initialise from the latest Solver:
\begin{equation}
    Q_t \;\leftarrow\; \text{GRPO}\!\bigl(S_{t-1},\; \text{vLLM}(S_{t-1})\bigr).
    \label{eq:prism_iter}
\end{equation}

Re-deriving the Questioner from the Solver at every iteration eliminates the capability lag inherent in the $Q_{t-1}\!\to\!Q_t$ chain, provides an unbiased starting point free of accumulated generation biases, and prevents the progressive distributional narrowing that arises when a policy is repeatedly tuned from its own already-narrowed predecessor~\cite{shumailov2023curse}.

\begin{algorithm}[t]
\caption{Prism: Coverage-Aware Self-Evolving Training}
\label{alg:prism}
\begin{algorithmic}[1]
\REQUIRE Base model $M_0$; fixed centroids $\{\boldsymbol{\mu}_k\}_{k=1}^{K}$; embedding model $f$; iterations $T$; diversity weight $\lambda$; EMA decay $\gamma$; smoothing constant $\alpha$
\ENSURE Trained Solver $S_T$
\STATE $Q_0 \leftarrow M_0$,\quad $S_0 \leftarrow M_0$,\quad $n_k \leftarrow \alpha \;\forall\, k \in [K]$
\FOR{$t = 1$ \TO $T$}
    \vspace{2pt}
    \STATE \textbf{// Questioner training}
    \STATE $Q_t \leftarrow S_{t-1}$ \COMMENT{initialise from latest Solver}
    \IF{$t > 1$}
        \STATE Warm-start $\mathbf{n}$ from counts at end of iteration $t{-}1$
    \ENDIF
    \FOR{each GRPO step}
        \STATE Sample question batch $\{q_g\}_{g=1}^{G} \sim \pi^Q_{Q_t}$
        \STATE Score each $q_g$: draw $n$ solver roll-outs, compute $p(q_g) = \frac{1}{n}\sum_i \mathsf{V}(q_g,a_i)$
        \FOR{each $q_g$ with $p(q_g) \in [0.5,\,0.9]$}
            \STATE $c(q_g) \leftarrow \arg\max_k\,\langle f(q_g),\boldsymbol{\mu}_k\rangle$ \COMMENT{cluster assignment}
            \STATE $d(q_g) \leftarrow \exp\!\bigl(-n_{c(q_g)}/\bar{n}\bigr)$ \COMMENT{rarity bonus}
            \STATE $r(q_g) \leftarrow \mathrm{zpd}(p(q_g))\cdot\bigl(1+\lambda\,d(q_g)\bigr)$ \COMMENT{final reward}
        \ENDFOR
        \STATE Update $Q_t$ via GRPO using rewards $\{r(q_g)\}$
        \STATE Apply EMA decay once: $n_k \leftarrow \gamma\,n_k\;\forall k$; then for each generated question update $n_{c(q_g)} \leftarrow n_{c(q_g)} + (1{-}\gamma)$ \COMMENT{EMA coverage update}
    \ENDFOR
    \vspace{2pt}
    \STATE \textbf{// Question generation \& Solver training}
\STATE Generate question pool $\mathcal{P}_t$ from $Q_t$; filter to $p(q)\in[0.5,0.9]$
    \STATE Train $S_t$ via GRPO on $\mathcal{P}_t$ with majority-vote rewards
\ENDFOR
\RETURN $S_T$
\end{algorithmic}
\end{algorithm}

\subsection{Solver Training}

Solver training follows R-Zero without modification. The Solver $S_t$ is trained via GRPO on questions generated by $Q_t$, with majority voting over $n{=}8$ roll-outs providing pseudo-labels and outcome-based rewards.
This deliberate choice keeps the solver-side optimisation procedure fixed, so performance differences between Prism and R-Zero are driven primarily by question-side changes (the generated curriculum distribution, and the exploration dynamics induced by the Questioner reward and initialisation).

\section{Experimental Setup}

\subsection{Models and Training}

All experiments use \textbf{Qwen3-4B-Base} as the shared initialisation for both the Questioner and Solver.
Our implementation builds on the \texttt{verl} framework~\cite{huang2025rzero} and is trained on an 8$\times$H100 node.
Both the R-Zero baseline and Prism are trained for $T{=}4$ co-evolution iterations under identical Solver training procedures; only the Questioner reward and initialisation differ.
The Questioner is trained via GRPO for 6 steps per iteration with 4 roll-outs, and the Solver for 20 steps with 8 roll-outs.
We use \texttt{Qwen3-Embedding-0.6B} as the embedding model for cluster assignment in Prism, with $K{=}128$ clusters, diversity weight $\lambda{=}5.0$, and EMA decay $\gamma{=}0.99$.
Full hyperparameters are listed in Appendix~\ref{app:hyperparams}.

\subsection{Baselines}

We compare Prism against five self-evolving or self-play baselines (counting R-Few at both 5\% and 1\% anchor settings), all using Qwen3-4B-Base as the shared initialisation:
\begin{itemize}[leftmargin=*,itemsep=2pt]
    \item \textbf{R-Zero}~\cite{huang2025rzero}: Data-free Challenger and Solver co-evolution with an uncertainty reward and BLEU-based repetition penalty; the Challenger at iteration $t$ is initialised from $Q_{t-1}$. We re-implement R-Zero under conditions identical to Prism (same base model, hardware, training budget, and Solver procedure) to ensure a fair comparison.
    \item \textbf{R-Few}~\cite{yu2025rfew}: Extends R-Zero by injecting a small pool of human-labelled anchor examples as in-context guidance. We report published numbers for both the 5\% and 1\% anchor pool sizes.
    \item \textbf{Absolute Zero}~\cite{zhao2025absolutezero}: A fully data-free self-play framework that jointly trains a proposer and solver with reinforced self-play, requiring zero external data.
    \item \textbf{SPICE}~\cite{chen2025spice}: Corpus-grounded self-play in which the model generates questions conditioned on retrieved passages and trains on its own outputs.
\end{itemize}
All methods share the same base model and evaluation protocol. For R-Zero we use our own re-implementation; for R-Few, Absolute Zero, and SPICE we report published numbers evaluated under the same benchmark suite.

\subsection{Evaluation}

We evaluate on seven mathematical reasoning benchmarks spanning a wide difficulty spectrum:
\textbf{GSM8K}~\cite{cobbe2021gsm8k} (grade-school), \textbf{MATH-500}~\cite{hendrycks2021math} (competition-level), \textbf{AMC}~\cite{maa_amc} (high-school competition), \textbf{Minerva Math}~\cite{lewkowycz2022minerva} (undergraduate), \textbf{OlympiadBench}~\cite{he2024olympiadbench} (olympiad-level), \textbf{AIME 2024}~\cite{aops_aime2024}, and \textbf{AIME 2025}~\cite{aops_aime2025} (competition math, integer answers).
This range is important because if curriculum diversity is the binding constraint, its impact should be most visible on the hardest tasks, where success demands the broadest coverage of problem-solving strategies.
All benchmarks are evaluated with Pass@1 accuracy.

\section{Results}

\subsection{Main Results}

Table~\ref{tab:main_results} reports Pass@1 accuracy across all seven benchmarks.
Prism achieves the highest accuracy on six of seven benchmarks, outperforming all five baselines, including R-Zero, R-Few (both 1\% and 5\%), Absolute Zero, and SPICE.
The gains are especially pronounced on harder evaluations, including $+3.37$ points over R-Zero on AIME~2024 (16.77 vs.\ 13.40), $+3.68$ on Minerva Math (56.62 vs.\ 52.94), and $+3.98$ on AMC (61.25 vs.\ 57.27).
On AIME~2025, SPICE achieves the highest score (19.10).

\begin{table}[h]
\centering
\setlength{\tabcolsep}{3pt}
\resizebox{\linewidth}{!}{%
\begin{tabular}{@{}lccccccc@{}}
\toprule
\textbf{Model} & \textbf{GSM8K} & \textbf{MATH-500} & \textbf{AMC} & \textbf{Minerva} & \textbf{OlyBench} & \textbf{AIME 2024} & \textbf{AIME 2025} \\
\midrule
Qwen3-4B-Base & 72.60 & 68.20 & 47.50 & 42.30 & 34.80 & 6.70 & 10.30 \\
\midrule
R-Zero~\cite{huang2025rzero} & 92.12 & 79.60 & 57.27 & 52.94 & 44.59 & 13.40 & 9.60 \\
R-Few (5\%)~\cite{yu2025rfew} & 92.60 & 78.00 & 52.40 & 53.20 & 42.80 & 14.50 & 9.90 \\
R-Few (1\%)~\cite{yu2025rfew} & 92.30 & 77.80 & 52.70 & 52.10 & 42.40 & 13.60 & 9.10 \\
Absolute Zero~\cite{zhao2025absolutezero} & 89.30 & 76.20 & 52.50 & 38.20 & 38.50 & 13.30 & 13.30 \\
SPICE~\cite{chen2025spice} & 92.70 & 78.00 & 57.50 & 51.90 & 42.70 & 12.20 & \textbf{19.10} \\
\midrule
Prism (ours) & \textbf{93.45} & \textbf{81.02} & \textbf{61.25} & \textbf{56.62} & \textbf{45.58} & \textbf{16.77} & 12.92 \\
\bottomrule
\end{tabular}%
}
\caption{Pass@1 accuracy across seven mathematical reasoning benchmarks. All methods use Qwen3-4B-Base as the base model. \textbf{Bold} indicates the best result per benchmark.}
\label{tab:main_results}
\end{table}

\subsection{Curriculum Diversity Analysis}
\label{sec:diversity_analysis}

To verify that solver gains arise from improved curriculum coverage, we embed questions generated by each questioner using \texttt{Qwen3-Embedding-0.6B} and compute per-cluster frequency vectors over the fixed $K{=}128$ partition.

\paragraph{Distributional statistics.}
Table~\ref{tab:coverage_stats} summarises coverage statistics for the base model, R-Zero, and Prism.
Self-evolution under R-Zero \emph{worsens} coverage relative to the base model. Active clusters drop from 89 to 65, the Gini coefficient rises to 0.90, and a single cluster (cluster~44) accumulates 951 questions, nearly 50$\times$ the mean.
Prism reverses this trend. With 107 of 128 clusters active, normalised entropy reaches 0.83 and the Gini falls to 0.66, even below the base model's 0.81.

\begin{table}[h]
\centering
\caption{Semantic coverage statistics over $K{=}128$ clusters. Higher entropy, lower Gini, and lower top-10 share indicate more uniform coverage.}
\label{tab:coverage_stats}
\small
\begin{tabular}{@{}lcccccc@{}}
\toprule
\textbf{Questioner} & \textbf{Active} & \textbf{Std} & \textbf{Entropy} & \textbf{Norm.\ Ent.} & \textbf{Gini} & \textbf{Top-10 \%} \\
\midrule
Base model          & 89  & 33.3 & 4.96 & 0.71 & 0.81 & 61.2\% \\
R-Zero              & 65  & 91.4 & 3.71 & 0.53 & 0.90 & 79.5\% \\
Prism               & \textbf{107} & \textbf{36.0} & \textbf{5.81} & \textbf{0.83} & \textbf{0.66} & \textbf{42.4\%} \\
\bottomrule
\end{tabular}
\end{table}

\begin{figure}[h]
\centering
\begin{minipage}[t]{0.65\linewidth}
    \centering
    \includegraphics[width=\linewidth]{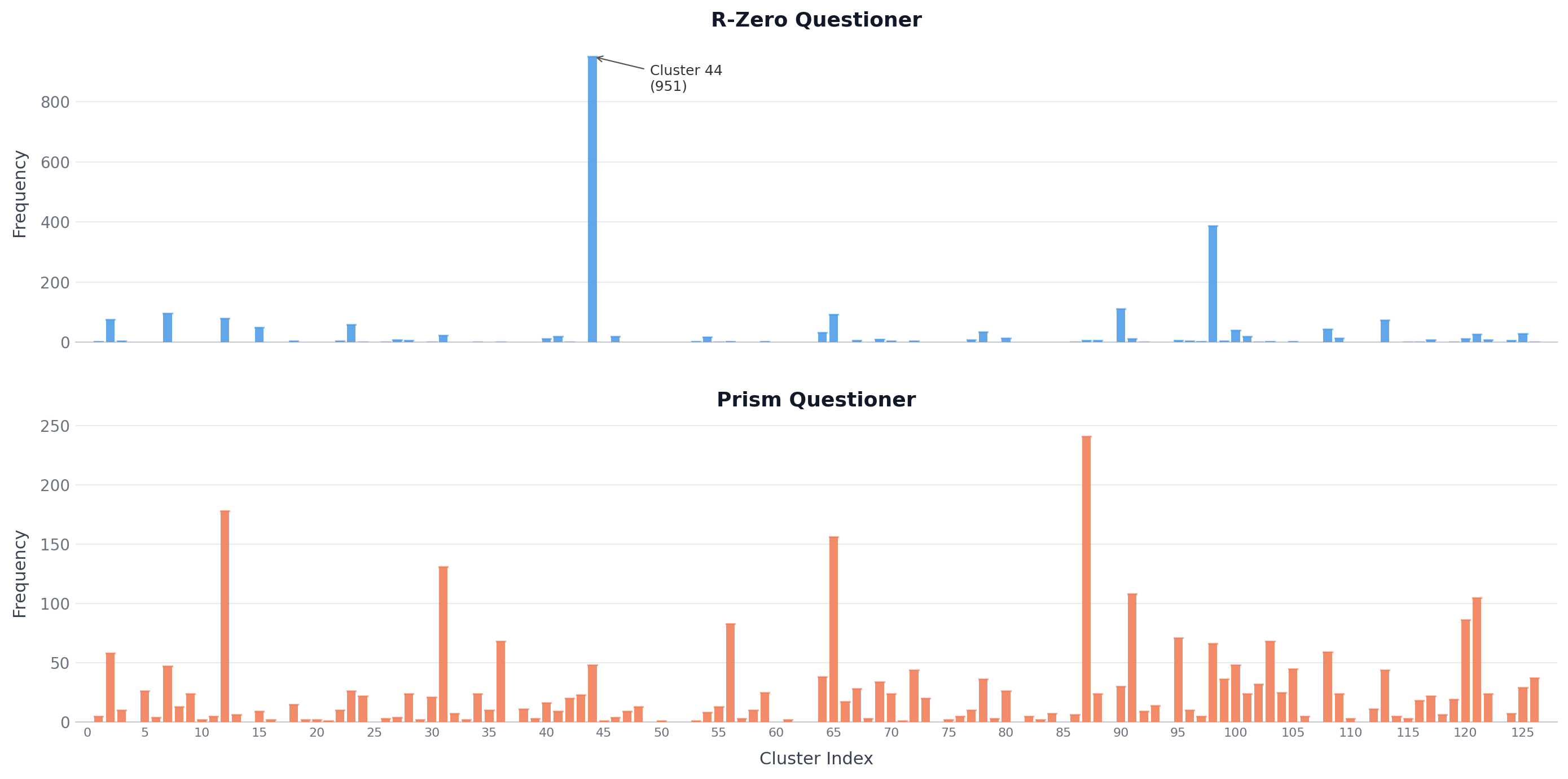}
\end{minipage}%
\hspace{0.02\linewidth}%
\begin{minipage}[t]{0.31\linewidth}
    \centering
    \includegraphics[width=\linewidth]{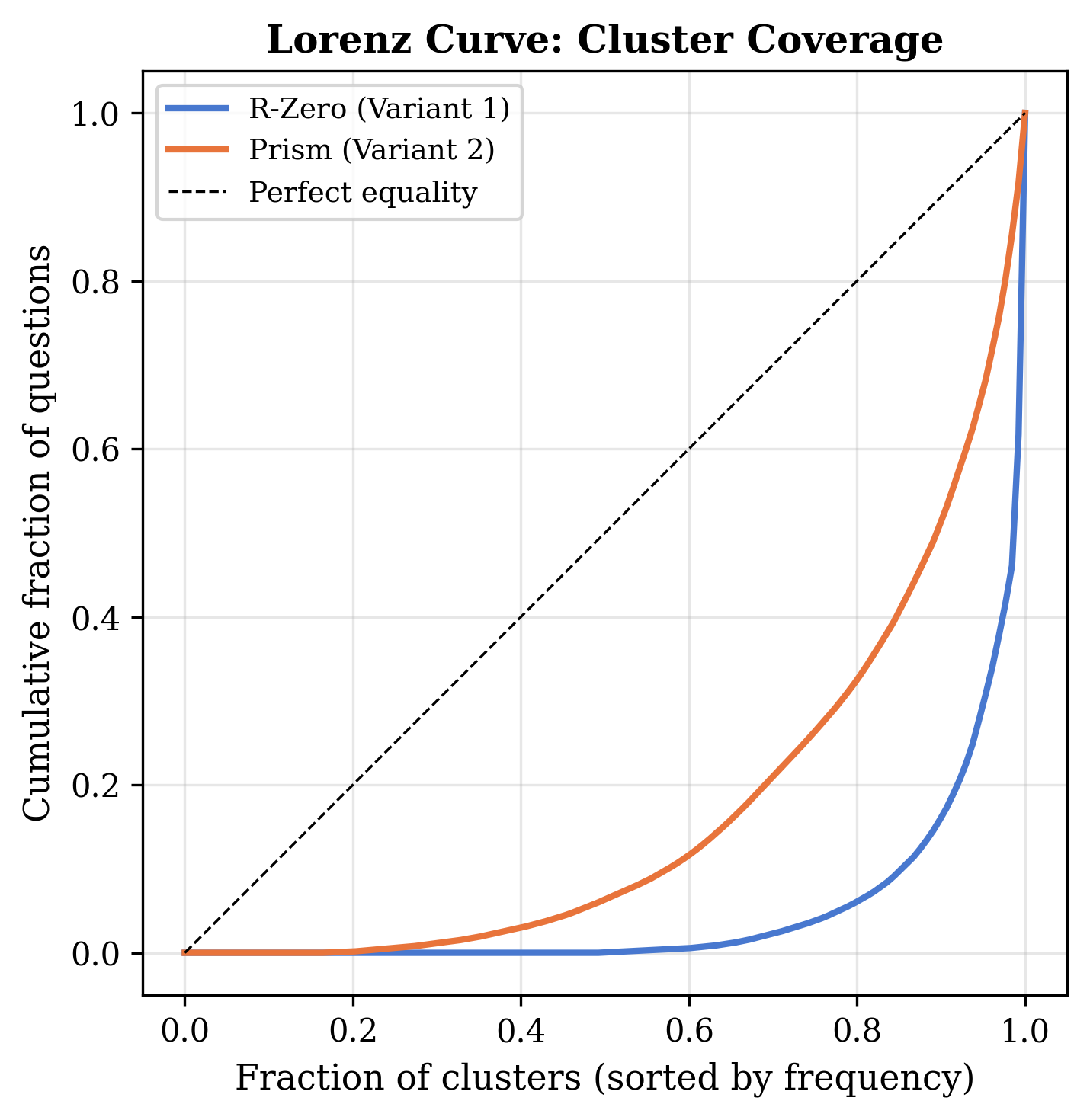}
\end{minipage}
\caption{Curriculum diversity visualisation. \textbf{Left:} Per-cluster frequency distributions. R-Zero (top) is dominated by a single spike at cluster~44 (951 questions); Prism (bottom) covers 107 of 128 clusters with no dominant mode. \textbf{Right:} Lorenz curves for R-Zero (blue) and Prism (orange). A curve closer to the diagonal indicates more equitable cluster coverage.}
\label{fig:diversity_vis}
\end{figure}

\paragraph{Visualisation.}
Figure~\ref{fig:diversity_vis} visualises the coverage gap.
The left panel shows per-cluster frequency bars; the R-Zero distribution is sharply spiked while Prism's is broadly distributed with no dominant cluster.
The right panel shows the corresponding Lorenz curves. The R-Zero curve bows well below the diagonal (the bottom 80\% of clusters hold $<$10\% of questions), while Prism lies substantially closer to it.

\paragraph{Qualitative evidence.}
Inspection of the generated questions confirms the statistical picture.
R-Zero's dominant cluster (cluster~44) consists almost entirely of structurally near-identical polynomial-divisibility problems, exemplifying the \emph{diversity illusion}~\cite{li2025rdiverse}.
Prism, by contrast, distributes generation across combinatorics, geometry, functional equations, and modular arithmetic, covering regions that R-Zero leaves entirely empty.
Representative examples from both systems are provided in Appendix~\ref{app:qualitative}.

\paragraph{Prism-Math dataset.}
\label{sec:prism_math}
As a byproduct of this process, we release \textbf{Prism-Math}, a dataset of ${\sim}$100K questions generated by the Prism questioner.
Unlike existing synthetic datasets, Prism-Math is explicitly optimised for semantic breadth and edge-of-solvability difficulty, making it well-suited for further training and topic-coverage evaluation.

\subsection{Ablation Studies}
\label{sec:ablation}

We conduct ablation experiments to isolate the contribution of each Prism component, study sensitivity to key hyperparameters, and quantify computational overhead.

\paragraph{Component ablation.}
Prism introduces two modifications over R-Zero: (a) a cluster-diversity reward and (b) Solver-initialised Questioner.
Table~\ref{tab:ablation} disentangles their effects using a $2{\times}2$ factorial design on MATH-500.

\begin{table}[h]
\centering
\small
\begin{tabular}{@{}cc|cc@{}}
\toprule
\textbf{Div.\ Reward} & \textbf{SolverInit} & \textbf{MATH-500}  \\
\midrule
\xmark & \xmark & 79.60 \\
\cmark & \xmark & 80.46 \\
\xmark & \cmark & 79.12 \\
\cmark & \cmark & \textbf{81.02} \\
\bottomrule
\end{tabular}
\caption{Component ablation (Pass@1 accuracy). ``Div.'' = cluster-diversity reward; ``SolverInit'' = Solver-initialised Questioner.}
\label{tab:ablation}
\end{table}

The cluster-diversity reward alone yields a $+0.86$ point gain over the R-Zero baseline (80.46 vs.\ 79.60), confirming that coverage pressure improves curriculum quality even under the standard $Q_{t-1}\!\to\!Q_t$ initialisation.
The Solver-initialised Questioner alone does not improve over the baseline ($-0.48$). This is expected: the Solver is directly optimised to maximise outcome rewards \emph{conditioned on a question} (i.e., to answer), so its generation prior is not automatically calibrated for proposing diverse, informative, edge-of-solvability questions. Without an explicit question-side objective (here, the coverage-aware reward), initialising the Questioner from the Solver can reduce the quality/learnability of generated questions, even if the underlying model is more capable at solving.
However, the full combination achieves the highest accuracy (81.02, $+1.42$ over the baseline). The Solver-derived initialisation provides an unbiased starting point from which the diversity reward can steer effectively, while the diversity reward supplies the exploration pressure that the neutral initialisation alone lacks.

\paragraph{Sensitivity to diversity weight $\lambda$.}
We vary $\lambda \in \{1.0, 3.0, 5.0, 8.0\}$ while holding $K{=}128$ fixed.

\begin{table}[h]
\centering

\label{tab:sensitivity_lambda}
\small
\begin{tabular}{@{}l|ccc@{}}
\toprule
$\lambda$ & \textbf{MATH-500} \\
\midrule
1.0 & 78.32 \\
3.0 & 79.80 \\
5.0 & \textbf{81.02} \\
8.0 & 78.80 \\
\bottomrule
\end{tabular}
\caption{Sensitivity to the diversity weight $\lambda$.}
\end{table}

At $\lambda{=}1.0$ the diversity pressure is too weak to overcome the Questioner's natural mode-seeking behaviour, while at $\lambda{=}8.0$ the rarity bonus dominates to the point where the Questioner begins to chase obscure clusters at the expense of generating high-quality, edge-of-solvability questions.
The sweet spot at $\lambda{=}5.0$ balances coverage and quality effectively.

\paragraph{Computational overhead.}
Prism introduces two additional per-batch operations: (i) embedding each generated question with \texttt{Qwen3-Embedding-0.6B} and (ii) computing cluster assignments and updating the EMA count vector.
Table~\ref{tab:overhead} reports wall-clock times per co-evolution iteration measured on our 8$\times$H100 node.

\begin{table}[h]
\centering
\label{tab:overhead}
\small
\begin{tabular}{@{}lcc@{}}
\toprule
\textbf{Component} & \textbf{R-Zero} & \textbf{Prism} \\
\midrule
Questioner GRPO training & 38 min & 38 min \\
Embedding + cluster assignment & N/A & 2.4 min \\
Question generation \& filtering & 25 min & 25 min \\
Solver GRPO training & 82 min & 82 min \\
\midrule
\textbf{Total per iteration} & 145 min & 147.4 min \\
\textbf{Overhead} & N/A & \textbf{+1.7\%} \\
\bottomrule
\end{tabular}
\caption{Wall-clock overhead per co-evolution iteration.}
\end{table}

The embedding and clustering step adds only ${\sim}$2.4 minutes ($+1.7\%$) per iteration, making Prism's diversity mechanism effectively free relative to the GRPO training cost.

\subsection{Limitations and Future Work}

The semantic partition is static. The $K{=}128$ clusters are built once from the MATH training set and cannot represent problem types outside the reference corpus, nor does the current design reward questions that \emph{compose} skills across clusters.
Replacing the fixed partition with an adaptive, online clustering scheme that splits or spawns niches as the Questioner's distribution evolves is a natural next step.
The partition quality also depends on the embedding model (\texttt{Qwen3-Embedding-0.6B}); we have not ablated across embedding architectures, and embeddings that conflate distinct problem types would silently weaken the diversity signal.
Since coverage is defined and measured in this same embedding space, additional evaluation with independent embedding models or non-embedding redundancy metrics would further strengthen the diversity analysis.
On the verification side, Prism uses outcome-only rewards (exact-match or symbolic equivalence), which can assign credit to correct answers derived through flawed reasoning; combining the coverage bonus with process-aware reward models could simultaneously encourage semantic breadth and derivation correctness.
Finally, due to resource constraints, all experiments use a single base model (Qwen3-4B-Base) on mathematical reasoning; extending Prism to other model scales and domains such as code generation or scientific reasoning remains open.

\section{Conclusion}

Self-evolving reasoning systems can exhibit \emph{curriculum collapse}, in which the Questioner concentrates on a narrow slice of the problem space as training progresses, reducing semantic coverage across iterations and limiting downstream Solver improvement.
We presented \textbf{Prism}, a question-centric self-evolution framework that explicitly regularises cross-iteration semantic coverage via an embedding-induced semantic partition and a coverage-aware reward, while preserving difficulty through ZPD gating and reducing inter-iteration drift by re-deriving the Questioner from the latest Solver.
Our experiments and analyses show that enforcing persistent coverage changes the training distribution in the intended direction, mitigating semantic mode collapse without disrupting the underlying RLVR loop.
We release the code, trained models, and Prism-Math to facilitate further research on coverage-regularised self-evolution.

More broadly, our results suggest that question generation should be treated as a first-class optimisation target in closed-loop RLVR, and that controlling the Questioner's cross-iteration coverage is a direct lever for building more capable self-evolving reasoners.

\bibliographystyle{plain}

\clearpage
\appendix

\begin{center}
    {\Large\bfseries Appendix}
\end{center}
\vspace{1em}

\section{Training Hyperparameters}
\label{app:hyperparams}

We list all training hyperparameters used for training.
All experiments are conducted using BFloat16 mixed precision and FlashAttention~2.

\begin{table}[h]
\centering

\label{tab:hyperparams}
\small
\begin{tabular}{@{}lll@{}}
\toprule
\textbf{Category} & \textbf{Setting} & \textbf{Value} \\
\midrule
\multirow{4}{*}{General}
& Base model & Qwen3-4B-Base \\
& Co-evolution iterations ($T$) & 4 \\
& Hardware & 8$\times$ H100 node \\
& Precision & BFloat16 \\
\midrule
\multirow{4}{*}{Questioner}
& GRPO steps per iteration & 6 \\
& Roll-outs ($n$) & 4 \\
& Initialisation at iter $t$ & $S_{t-1}$ (Prism) / $Q_{t-1}$ (R-Zero) \\
& Learning rate & $5 \times 10^{-6}$ \\
\midrule
\multirow{5}{*}{Solver}
& GRPO steps per iteration & 20 \\
& Roll-outs ($n$) & 8 \\
& KL penalty coefficient & $1 \times 10^{-4}$ \\
& Learning rate & $5 \times 10^{-6}$ \\
\midrule
\multirow{5}{*}{Prism-specific}
& Cluster count ($K$) & 128 \\
& Embedding model & Qwen3-Embedding-0.6B \\
& Diversity weight ($\lambda$) & 5.0 \\
& EMA decay ($\gamma$) & 0.99 \\
& Smoothing constant ($\alpha$) & 1.0 \\
\midrule
\multirow{2}{*}{Evaluation}
& Majority-vote threshold & 0.5 \\
& Decoding & Greedy (Pass@1) \\
\bottomrule
\end{tabular}
\caption{Training hyperparameters for Prism.}
\end{table}

\section{Qualitative Analysis of Generated Questions}
\label{app:qualitative}

To qualitatively illustrate curriculum collapse versus semantic breadth, we provide an example set of questions sampled from our training of R-Zero and Prism.

\paragraph{R-Zero: near-duplicate template (polynomial divisibility).}
The following five R-Zero questions are structurally almost identical: they repeatedly instantiate the same ``find the smallest $n$ such that a polynomial $P(x)$ is divisible by $Q(x)$'' template, with only minor changes in $Q(x)$ (and occasionally $P(x)$):
\begin{enumerate}[leftmargin=*,itemsep=2pt]
    \item Find the smallest positive integer \( n \) such that the polynomial \( P(x) = x^n + x^{n-1} + \cdots + x + 1 \) is divisible by the polynomial \( Q(x) = x^3 + x^2 + x + 1 \).
    \item Find the smallest positive integer \( n \) such that the polynomial \( P(x) = x^n + x^{n-1} + \cdots + x + 1 \) is divisible by \( Q(x) = x^4 + 1 \).
    \item Find the smallest positive integer \( n \) such that the polynomial \( P(x) = x^n + x^{n-1} + \cdots + x + 1 \) is divisible by the polynomial \( Q(x) = x^2 + x + 1 \).
    \item What is the smallest positive integer \( n \) such that the polynomial \( P(x) = x^n + 1 \) is divisible by the polynomial \( Q(x) = x^2 + x + 1 \)?
    \item Determine the smallest positive integer \( n \) such that the polynomial \( P(x) = x^n + x^{n-1} + \cdots + x + 1 \) is divisible by \( Q(x) = x^2 + x + 1 \) for some integer \( n > 1 \). Provide your answer modulo 1000.
\end{enumerate}

\paragraph{Prism: diverse topics and forms.}
In contrast, the following five Prism questions span geometry, combinatorics, algebra, modular arithmetic, and optimisation, with no dominant shared template beyond being mathematical reasoning problems:
\begin{enumerate}[leftmargin=*,itemsep=2pt]
    \item In triangle \( ABC \), side \( AB \) is extended beyond \( B \) to \( D \) so that \( BD = AB \), side \( BC \) is extended beyond \( C \) to \( E \) so that \( CE = BC \), and side \( CA \) is extended beyond \( A \) to \( F \) so that \( AF = CA \). If the area of \( \triangle ABC \) is 1, find the area of \( \triangle DEF \).
    \item Find the number of ways to arrange the letters in the word ``MATHEMATICS'' such that no two vowels are adjacent to each other.
    \item Solve for \( y \) in terms of \( x \): \(\; \frac{x + y}{3} - \frac{x - y}{2} = 4 \).
    \item What is the remainder when \(7^{2003}\) is divided by 25?
    \item In a magical land, there are two types of coins: golden coins worth 3 units and silver coins worth 7 units. A wizard has a chest containing a combination of these coins such that the total value of the coins is exactly 100 units. What is the minimum number of coins the wizard must have in his chest?
\end{enumerate}

\end{document}